\documentclass{article}

\usepackage{Odyssey2016,epsfig,amssymb,amsmath}
\usepackage{amssymb}
\usepackage{graphicx}
\usepackage{subfigure}
\usepackage{algorithm}
\usepackage{algorithmic}
\usepackage{amsmath}
\usepackage{amscd}
\usepackage{amsbsy}
\usepackage{hyperref}
\usepackage{url}
\usepackage{amsmath}
\usepackage{tikz}
\usepackage{subfigure}
\usetikzlibrary{arrows}
\usetikzlibrary{matrix}

\newcommand{\T}{{\scriptscriptstyle \top}}
\newcommand{\comment}[1]{}

\title{ A Semisupervised Approach for Language Identification based on Ladder Networks}

\name  {Ehud Ben-Reuven  \,\,  and   \,\, Jacob Goldberger}
\address{  Engineering Faculty \\  Bar-Ilan University, Israel \\
  {\small \tt udi@benreuven.com \hspace{1cm}  jacob.goldberger@biu.ac.il}    }

\begin{document}

    \maketitle

\sloppy

\begin{abstract}
In this study we address the problem of
 training a neural-network for language identification  using both labeled and unlabeled speech samples
in the form of i-vectors.
We propose a neural network architecture that can also handle out-of-set languages.
  We utilize a modified version of the recently proposed Ladder Network semisupervised training procedure that optimizes the reconstruction costs of a stack of denoising autoencoders.   We show that this approach can be  successfully  applied  to the case where the training dataset is composed of both labeled and unlabeled acoustic data.   The results show enhanced  language identification on the  NIST 2015 language identification dataset.
\end{abstract}

\section{Introduction}

The purpose of automatic language recognition is to identify which language is spoken
from a speech sample.  There are many characteristics of speech that could be used to identify languages.
Languages are made up of different sounds that form phonemes, so it is  possible to distinguish languages based on the acoustic features present in the speech signal. There are of course also lexical information. Languages are separable by the vocabulary, or sets of words and syntactic rules.
In this study we focus on language identification that is  based solely on the acoustic information conveyed by the speech signal.
 The applications of language identification systems include multilingual
translation systems and emergency call routing,  where the response time of a fluent native operator might be critical.

The impressive performance improvement obtained using deep
neural networks (DNNs) for automatic speech recognition
(ASR) \cite{Hinton2012} have motivated the application of DNNs to speaker and language
recognition.  DNN have been  trained for a different purpose to learn frame-level features that were  then used to train
a secondary classifier for the intended language recognition task
\cite{Yong2013}\cite{Matejka2014}. DNNs have also been  applied to directly train language classification systems  \cite{Lopez2014}\cite{Richardson2015}.

In this study we applied DNN to language recognition and report performance on the NIST  2015 Language Recognition i-vector Machine Learning Challenge \cite{NIST2015}.
In this task there are i-vectors examples from 50 languages. There are also unlabeled training data and we also need to address the problem of out-of-set languages that can appear in both the unlabeled training data and the test set. In this study we propose a DNN based approach that addresses both issues of  unlabeled training data and out-of-set examples.
Most previous DNN learning approaches have used  unlabeled data only for pre-training  which is a way to initialize the weights of the network
followed by normal supervised learning. The standard pre-training  strategy is based on a greedy layer-wise  procedure using either Restricted Boltzmann  Machines (RBM) \cite{Hinton2006} or noisy autoencoders \cite{Bengio2007}. In contrast, here we aimed   to explicitly incorporate the unlabeled data into the cost-function that is optimized in the training step.

Our approach is based on the recently proposed Ladder Network training procedure that has proven to be very successful,  especially in cases where the training dataset is composed of both labeled and unlabeled data \cite{1507.02672} \cite{1411.7783}. In addition to the supervised objective, the Ladder Network also has an unsupervised objective corresponding to the reconstruction costs of a stack of denoising autoencoders.
  However, instead of reconstructing one layer at a time and greedily stacking, an unsupervised objective involving the reconstruction of all layers is optimized jointly by all parameters (with the relative importance of each layer cost controlled by hyper-parameters).

  The paper proceeds as follows. In the next section we present a brief overview of Ladder Networks. We then describe a semisupervised language identification system based on Ladder Networks.  We demonstrate the  performance improvement of the proposed method  on a dataset from the NIST 2015 Language Recognition i-vector Challenge \cite{NIST2015}.

\begin{figure*}[thpb]
\center
\includegraphics[trim=0 420 0 50, clip, scale=0.8]{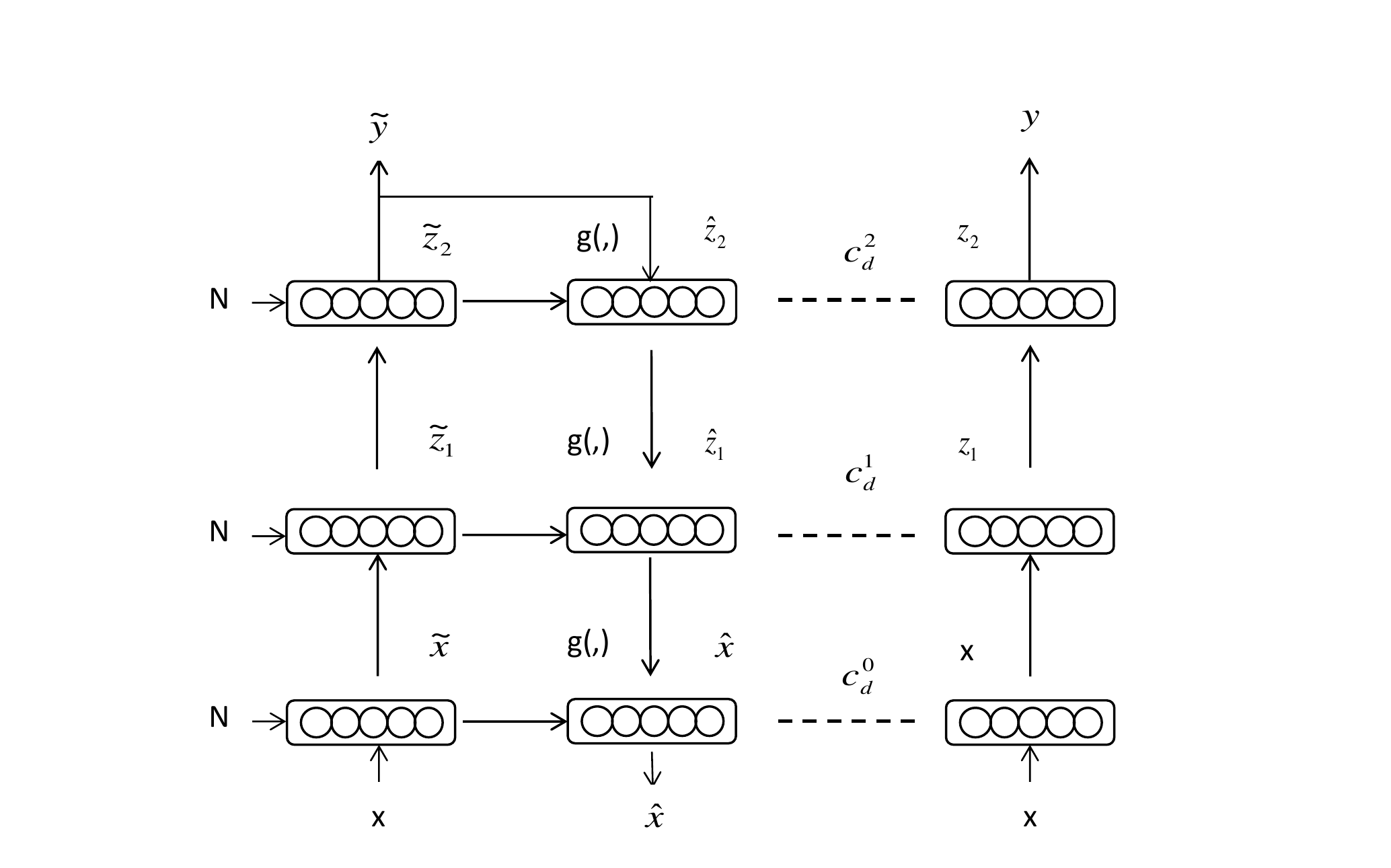}
\caption{A conceptual illustration of the Ladder Network for $L=2$ \cite{1507.02672}.}
\label{Ladder}
\end{figure*}

\section{The Ladder Network Architecture}

In this section we give a brief overview of the Ladder Network training technique. A detailed description of Ladder Networks can be found in \cite{1507.02672} \cite{1411.7783}. Assume we are given training data consisting of $n$ labeled examples  $x_1,...,x_n\in R^d$ with labels $c_1,...,c_n \in \{1,...,k\}$ and $m\!-\!n$ additional  unlabeled examples $x_{n\!+\!1},...,x_{m}$. We assume that the unrevealed labels belong  to the same set  $\{1,...,k\}$.

Our goal is to train a fully connected multi-layer  DNN using  both the labeled and the unlabeled training data.
Each layer of the network is formalized as a linear transformation followed by a non-linear activation function $\phi$:
\begin{equation}
\begin{aligned}
h_0 &= x \\
z_l &= W_l h_{l-1}, \hspace{1.0cm} l=1,...,L \\
h_l &= \phi (z_l)
\end{aligned}
\end{equation}
Softmax is used as the non-linear activation function in the last layer to compute the output class distribution $h_L= p(y|x)$.
In the Ladder Network training procedure in addition to the standard forward pass, we also apply a corrupted forward pass which is implemented by adding isotropic Gaussian noise to the input and to all hidden layers:
\begin{equation}
\begin{aligned}
 \widetilde{h}_0 &= \widetilde{x} =  x + \mbox{noise}\\
\widetilde{z}_l &= W_l \widetilde{h}_{l-1} + \mbox{noise}, \hspace{1.0cm} l=1,...,L\\
\widetilde{h}_l &= \phi (\widetilde{z}_l)\\
\end{aligned}
\end{equation}
Denote the soft-max output obtained by the noisy-pass by $\widetilde{y}$.
We view the noisy feed-forward pass as a data-encoding procedure that is followed by a denoising decoding pass whose
  target is the result of the noise-less forward propagation at each layer. The decoding procedure is defined as follows:
\begin{equation}
\begin{aligned}
u_L &=  \widetilde{h}_L\\
\hat{z}_L &= g  (\widetilde{z}_L, u_L) \\
u_l &=  V_{l\!+\!1} \hat{z}_{l\!+\!1} , \hspace{1.5cm} l=L\!-\!1,...,0\\
\hat{z}_l &= g  (\widetilde{z}_l, u_l) \\
\hat{x} &= \hat{z}_0
\end{aligned}
\end{equation}
where $V_l$ is a weight matrix from layer $l$ to layer $l\!-\!1$, with the same dimensions of $W_l^{^\T}$.  The function $g(\cdot, \cdot)$ is called the combinator
function as it combines the vertical $u_{l}$ and the lateral $\widetilde{z}_l$ connections
(also called skip connections) in an element-wise fashion in order  to estimate $z_l$. The estimation is linear as a function of $\widetilde{z}_l$  whose coefficients are non-linear functions of  $u_{l}$. A comparison of other possible choices for the combinator function is presented in \cite{1511.06430}.
 The resulting encoder/decoder architecture thus resembles a ladder (hence the name Ladder Networks). Figure \ref{Ladder} illustrates a ladder structure with two hidden layers.

 The concept of batch normalization was recently proposed by Ioffe and Szegedy \cite{Ioffe2015} as a method to improve convergence as a result of reduced covariance shift. In batch normalization, we compute the running mean and variance of hidden units over the batch, and use them to ``whiten" hidden units in each mini-batch.  In Ladder Networks batch normalization plays another crucial role of preventing encoder and decoder collapse  to a trivial solution of a constant function  that can be easily denoised.  Note that for a standard autoencoder this cannot happen since we cannot change the input feature vector we want to reconstruct.  To simplify the presentation of the Ladder Network we ignored the batch normalization step that is incorporated in each layer of the Ladder Network.

\comment{
\begin{figure}[thpb    ]
\includegraphics[scale=0.28]{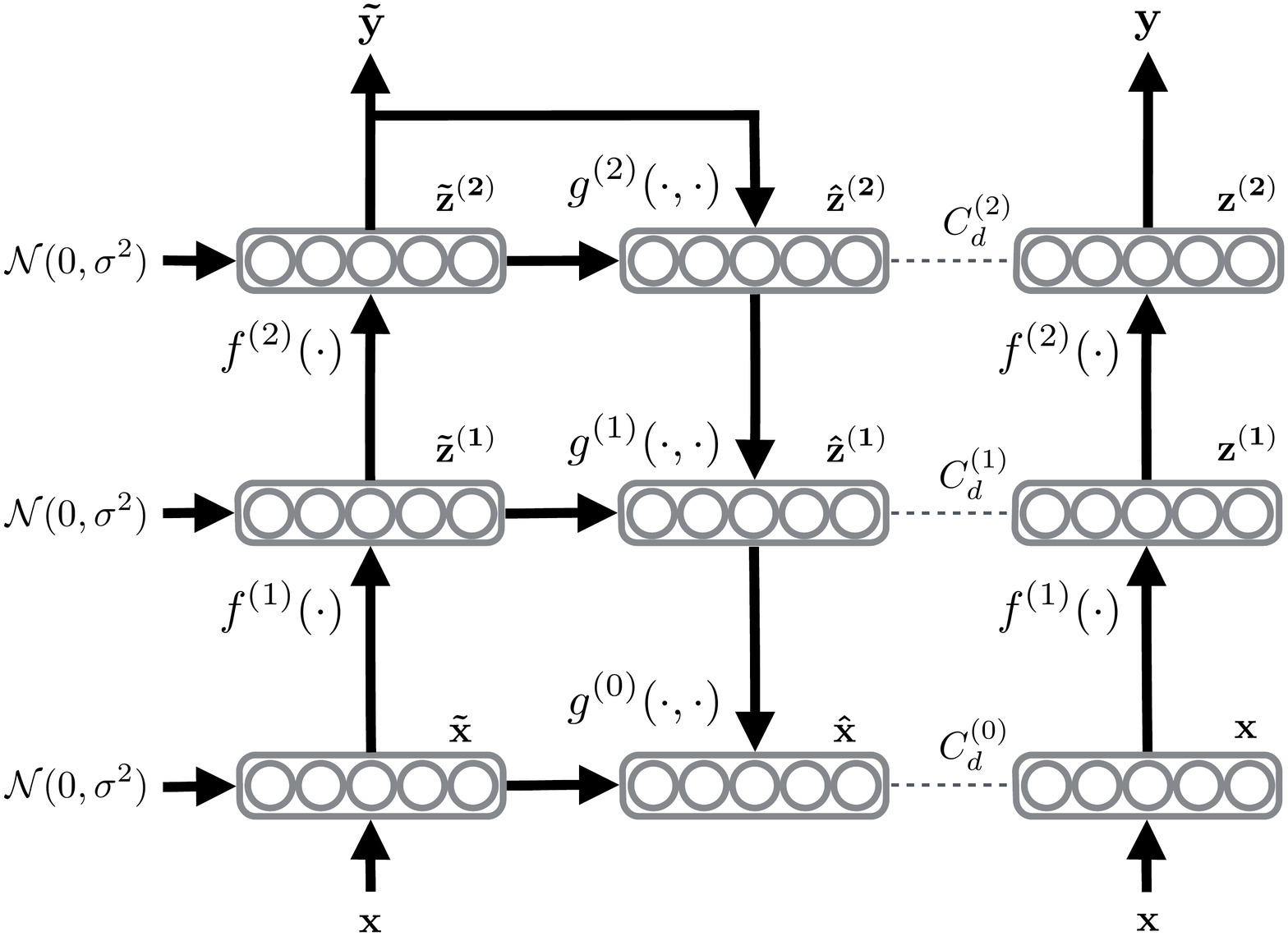}
\caption{A conceptual illustration of the Ladder Network for $L=2$ \cite{1507.02672}.}
\label{Ladder}
\end{figure}}
The cost function for training the network is composed of a supervised objective and an unsupervised autoencoder objective.
The supervised cost $C_s$ is the standard likelihood function which is defined as the average negative log probability of the
noisy output $\tilde y$ matching the target $c_t$ given the input $x_t$:
$$C_s = - \frac 1 n \sum_{t=1}^n \log p(\tilde y=c_t | x_t)$$
Note that, unlike standard DNN, the likelihood function is computed here based on the noisy network (and hence the back-propagation procedure is applied to the noisy feed-forward step). The  noise is used here to regularize supervised learning in a way similar to the weight noise regularization method \cite{Graves2011} and dropout \cite{srivastava2014}.

The cost of the standard denoising autoencoder is the input reconstruction error \cite{Vincent2008} \cite{Hinton2006a}. Here
the autoencoder cost function is the reconstruction error of the input along with all the network layers.
The error here is defined as the distance of the denoised decoded value from the noise-less forward value at each layer.
The unsupervised ladder cost function $C_d$ is:
\begin{equation}
C_d = \sum_{l=0}^L \lambda_l C_d^{(l)}
= \frac{1}{m} \sum_{l=0}^L \frac {\lambda_l} {w_l} \sum_{t=1}^m \left\Vert z_{l,t} - \hat{z}_{l,t} \right\Vert^2
\label{eq:cost}
\end{equation}
where $w_l$ is the layer's width, $m$ the number of training samples, and the hyperparameter
$\lambda_l$ is a layer-wise multiplier determining the importance of the denoising cost.
$z_{l,t}$ is the noise-free hidden layer obtained from the input $x_t$ and $\hat{z}_{l,t}$ is the corresponding decoded layer.
The target of the encode/decode at each layer is the result of the noise-less forward propagation.
Note that the ladder-autoencoder cost function does not use the labels and therefore it can be applied to both labeled and unlabeled training data.

A recent study \cite{1511.06430} conducted an extensive experimental investigation of variants of the Ladder Network in which  individual components were removed or replaced to gain insights into their relative importance. The authors found that all of the components were necessary for  optimal performance.  Ladder Networks have been found to obtain state-of-the-art results on standard datasets with only a small portion of labeled examples
 \cite{1507.02672}. This confirms a current trend in Deep Learning  that, while recent progress in DL  (e.g. \cite{Ioffe2015} \cite{srivastava2014}) applied on large labeled datasets does not rely on any unsupervised pretraining,  semisupervised learning might instead be crucial for success when only a small amount of labeled data is available.

\section{Dataset and Evaluation Metrics}
The 2015 language recognition i-vector challenge \cite{NIST2015} covers 50 target languages and a set of unnamed ``out-of-set" (oos) languages. Labeled training data (300 speech segments per language) were provided for the target languages and a set of approximately 6,500 unlabeled examples covering the target and out-of-set languages was provided for development. The test set consisted of 6,500 test segments covering the target and out-of-set languages. The speech duration of the audio segments used to create the i-vectors for the challenge were sampled from a log-normal distribution with a mean of approximately 35s.
The speech segments are derived from conversational telephone and narrow-band broadcast speech data.
Each speech segment is represented by an i-vector of 400 components \cite{Dehak}.

The task consists of classifying  each speech segment as either one of the  50 target languages or as an out-of-set.
 According to the challenge rules \cite{NIST2015rules} the goal is to minimize the following cost function:
\begin{equation} \mbox{Cost} = \frac{1\!-\!p_{\rm{oos}}}{k} \cdot \sum_{i=1}^k p_{\footnotesize \mbox{error}}(i) +
p_{\rm{oos}} \cdot p_{\footnotesize \mbox{error}}(\rm{oos})
\label{evalcost}
\end{equation}
where $k=50$, $p_{oos}=0.23$ is (assumed to be) the fraction of out-of-set examples in the test set and
$$p_{\footnotesize  \mbox{error}}(i) =  \frac{\mbox{\# errors-class-i}}{\mbox{\# trials-class-i}}   $$
is the fraction of examples in the test set with label $i$ that were misclassified.

\section{Model and Training}

\subsection{Network and cost function}

 Assume we are given training data consisting of $n$ labeled examples  $x_1,...,x_n\in R^d$ with labels $c_1,...,c_n \in \{1,...,k\}$. We also have a development set $x_{n\!+\!1},...,x_{m}$ with unspecified labels.  These unspecified labels include all  $k$ target labels and multiple additional out-of-set (oos) labels. Since we need to take care of the oos labels, we construct a DNN whose  soft-max output layer has  $k\!+\!1$ outputs corresponding to the  $k$ labels and to an out-of-set decision.
 Note that in order to train a  network with an oos explicit decision, we cannot rely only on  in-set labeled data.

\tikzstyle{int}=[draw, fill=blue!20, minimum size=2em]
\tikzstyle{init} = [pin edge={to-,thin,black}]

\begin{center}
\begin{tikzpicture}[node distance=3cm,auto,>=latex']

    \node [int, pin={[init]above:$x$}] (x) {$h_L$};


     \tikzstyle{neuron}=[circle,fill=black!25,minimum size=17pt,inner sep=0pt]

     \tikzstyle{output neuron}=[neuron, fill=blue!20];

    \foreach \name / \y in {1,...,6}
       \node[output neuron, below of=x] (I-\name ) at (3.5-\y,0) {};

      \node[output neuron, below of=x] (I-1 ) at (3.5-1,0) {oos};
    \node[output neuron, below of=x] (I-2 ) at (3.5-2,0) {$k$};
    \node[output neuron, below of=x] (1-6 ) at (3.5-6,0) {1};

        \path[->] (x) edge node {} (I-1);
        \path[->] (x) edge node {} (I-2);
        \path[dashed] (x) edge node {} (I-3);
        \path[dashed] (x) edge node {} (I-4);
        \path[dashed] (x) edge node {} (I-5);
        \path[->] (x) edge node {} (I-6);
\end{tikzpicture}
\end{center}

We applied  the  Ladder Network training scheme to take maximum advantage of the information conveyed in the unlabeled data. We first used the same ladder reconstruction error cost function that was described in the previous section. We used additional  cost function that is composed of two components.
The first score  is the negative likelihood score of the labeled data:
\begin{equation}
C_1 = - \frac 1 n \sum_{t=1}^n \log p(\tilde y=c_t | x_t).
\label{c1}
\end{equation}
where $\tilde y$ is the soft-output of the noisy network.
Denote the percentage of oos examples in the unlabeled development set (and in the test set) by $p_{oos}$. We assumed that $p_{oos}$ is given. We further assumed that there is an equal number of examples from each one of $k$ classes in the  unlabeled part of the training set   (and in the test set).
Define the estimated average label frequency  over all the unlabeled set as follows:
$$p_{av}(i) =  \frac 1 m \sum_{t=n+1}^{n+m}  p(\tilde y=i | x_t), \hspace{0.5cm} i=1,...,k $$
and
$$p_{av}(oos) =   \frac 1 m \sum_{t=n+1}^{n+m}  p(\tilde y=oos | x_t). $$
The second score encourages the estimated labels over all the development set to maintain the  (assumed to be) known label distribution.
The score is defined as follows:
\begin{equation}
C_2= -p_{oos} \log p_{av}(oos) - \frac{1\!-\!p_{oos}}{k} \sum_{i=1}^k  \log p_{av}(i).
\label{c2}
\end{equation}
In practice when we optimized the network parameter using mini-batches, we computed this score on each mini-batch instead of the entire development set. We set the size of the mini-batch to  be 1024 which is large enough in order to assume, based on the law of large numbers, that the label frequency within the mini-batch is similar to the overall label frequency.
The cost function $C_2$ is not part of the original ladder framework. It represents the assumption that the labeled training data and the unlabeled development set have similar label statistics. It also assumes a predefined relative number of oos  examples.
We show in the experiment section that, in addition to the contribution of the ladder method, this cost function further improves classification performance.

 The cost function of a batch is made up of the two elements described above:
\begin{equation}
C = C_1 + \alpha C_2
\label{traincost}
\end{equation}
where \(\alpha\) is an hyper-parameter that is tuned in a cross-validation procedure that is described below. (We can assume without loss of generality that one of the coefficients of the likelihood score, the label frequency score and the ladder score is 1.)

We tried to use another standard unsupervised cost function that is based on minimizing the entropy of the distribution obtained by the soft-max
layer, thus encouraging the decision in the unlabeled data to focus on one of the languages \cite{Grandvalet}:
$$
C_{ent} =  -\frac 1 m \sum_{t=n+1}^{n+m} \sum_i    p(\tilde y=i | x_t)  \log p(\tilde y=i | x_t)
$$
such that $i\in \{1,...,k\}\cup \{oos\}$.
 We tried computing this score with different importance weights either on the noise-free network or the noisy network.
  However, there was not any  significant  performance gain using this cost function.

    \subsection{A post processing step}\label{postprocessing}
The soft-max layer of the network provides a distribution over the 51 possible outputs.
 The final decision is then obtained by selecting the most probable label.
  Once the predictions are made on the test data, we can measure the ratio of the number of examples for which oos was predicted.
 If this ratio is below the pre-defined ratio, $p_{oos}$, we can relabel  more examples as oos until the
 expected ratio is met.  The candidate examples for relabeling are the ones in which probability of the most probable language is the smallest.
  If this ratio is above the pre-defined ratio, $p_{oos}$, we can relabel  fewer examples as oos until the
 expected ratio is met. This relabeling is done by multiplying the predicted probability to be oos by a reduction bias which gives other labels a higher likelihood of being picked as the predicted label.
  In the experiment results we show that this procedure indeed helps when using standard network training based only on the labeled data (i.e. when $\alpha\!=\!0$).   However,  there was no  improvement when this post-processing was applied on the Ladder Network results or when $\alpha\!>\!0$ in the standard network.

    \subsection{Parameter Tuning}\label{cross-validation}
We created a modified dataset for hyper parameters tuning using only the labeled part of the training dataset.
 We randomly selected 38 out 50 languages as the language-set and considered the data of the remaining 12 languages as out-of-set. We took a subset of the $38\!\times\!300$ examples as the labeled training set. We then used the remaining labeled data to construct a  development set and a test set that both  contained examples from all the 50 languages. The development and test sets were constructed
in a way that all the 50 languages (either in-set or out-of-set) were in the same
proportion.

The process of creating a modified set can be repeated many times, and each
time we chose another set of languages to be out-of-set,
thus giving each language an opportunity to be out-of-set.

The cross-validation described above was used to set the hyper-parameters of the Ladder Network.  The main parameters were the number of layers,  size of
each layer and  the relative importance of each layer in the reconstruction score. We also tuned the relative weights of  the score component $C_1$ and $C_2$.

\begin{table}
\caption{Progress-set scores for different models (pp stands for post-processing).}
\center
\begin{tabular}{|c|r|c|c|c|}
\hline
Model & $\alpha$ & score & oos ratio & pp \\ \hline
 baseline & 0 &  32.667 &  0.09 & 29.128 \\
 & 0.15 &  31.538 &  0.18 & 31.795 \\
\hline
 Ladder & 0 &  28.051 &  0.18  & 28.103 \\
 & 0.15 &  \bf{23.949} &  0.31 & 25.026 \\
\hline
\end{tabular}
\label{postable}
\end{table}

\begin{table}
\caption{Performance comparison of the Ladder and baseline  network on test subset for several partitions of the 50 languages into 38 in-set and 12 out-of set.}
\center
\begin{tabular}{|c|cc|cc|}
\hline
 & Ladder & \#epocs & baseline & \#epocs \\ \hline
1 & 19.1 &  680 &  24.0&  74 \\
2 & 18.3 & 887 & 22.8 & 112\\
3 & 20.9 & 961 & 29.1&  99\\
4 & 20.7 & 839 & 27.2 &  71\\
5 & 16.4 & 772 & 23.7 & 54\\
  \hline
  av & 19.1 &  & 25.3 & \\
  \hline
\end{tabular}
\label{restable}
\end{table}

\subsection{Model Configuration}

The network we implemented\footnote{code available at https://github.com/udibr/LRE} has an i-vector  input of dimension 400 which
passes through four ReLU hidden layers of size 500, 500,
500 and 100 and a final soft-max output layer of 51  (or 39 when doing cross validation for parameter tuning).
In the encoding step a  Gaussian noise with standard deviation of $\sigma=0.5$ was added to the input data
and to  all intermediate hidden layers.
Training was done with mini-batches having a  batch size of 1024.
Note that this size has a secondary effect through the loss function $C_2$ which
averages the  predictions  across the mini-batch  before computing the loss.
The training consisted of 1000  epoch iterations.
The order of the samples was shuffled before each iteration.
  It turned out that because of the unsupervised learning the ladder method is
insensitive to the number of epochs and having between 800 to 2000 epoch iterations gave
similar results.
A direct skip of information from the encoder to
the decoder was used only on the input layer
using the Gaussian method described in \cite{1507.02672}.
The $L_2$ reconstruction error of the denoising layers compared
with the noise-free network layers was assigned a weight of 1 for the input layer and the first
hidden layer and a weight of 0.3 for all other layers. The weights of $C_1$ and $C_2$  were set to 1 and 0.15 respectively.

\begin{figure}[hpb]
\center
\includegraphics[scale=0.75]{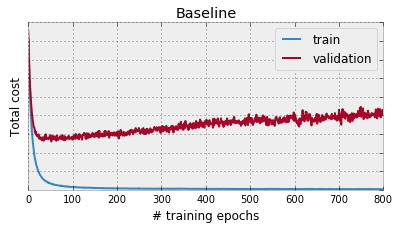}
\includegraphics[scale=0.75]{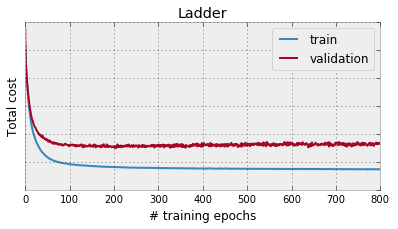}
\caption{Cost function score as a function of the number of epochs on the cross validation dataset.}
\label{fig_Ladder_score}
\end{figure}

\section{Experiments}

\subsection{NIST challenge results}

We next report the results for several variants of the method described above on the
 NIST 2015 Language Recognition i-vector Challenge \cite{NIST2015}.
Each combination of models and parameters was applied to the test set of the challenge and submitted to the competition web-site.
According to the challenge rules, an unknown subset of 30\% of the test samples was used to compute a score for the
progress-set and the results for the remaining 70\%  set are not reported by
the web-site.
The performance score is defined in Eq. (\ref{evalcost}) where a smaller number is better.
 Table \ref{postable} shows the progress-set results we achieved for
different models and parameters on the competition web site.

We examined two training procedures:  one with the full ladder structure and
one with a baseline variant where we set the weight of the ladder reconstruction score
to zero. The baseline method is still based on a noisy-forward step which, without the layer-reconstruction
cost function, turned out to be a  regularized learning method similar to dropout \cite{srivastava2014}.

For each model (either baseline or Ladder) we tested the results of using two variants of the training cost function (\ref{traincost}).
One variant (defined by $\alpha\!=\!0$) was based only on the likelihood score (\ref{c1}) computed on the labeled data.
In the second variant (defined by $\alpha\!=\!0.15$) we also used the cross entropy of the average probabilities  (\ref{c2}) computed on the unlabeled data. The optimal value of $\alpha$ was chosen using the cross-validation procedure on the labeled data that was described above.
Note that the Ladder Network uses the unlabeled data for training even if $\alpha=0$.
In contrast, the baseline method ignores the unlabeled data if   $\alpha=0$.
Table \ref{postable} shows that it is indeed beneficial to use  unlabeled data during  training as well.
The best results were obtained by Ladder training applied to the  cost function which besides a likelihood term  also used the cross entropy of the average probabilities. Table \ref{postable} reveals the importance of the cost function propose in this study  (\ref{c2}) that combined with the
Ladder network training method yielded the best results.

We next examined the contribution of the post-processing step described in  Section \ref{postprocessing}.
Table \ref{postable} also shows the ratio of oos decisions in the test data submitted to the challenge site.
We tried to improve the results by using the post-processing step (\ref{postprocessing})
with a parameter, $p_{oos}=0.23$, for the oos ratio. The results are shown in the rightmost column of Table \ref{postable}.
The post-processing step only led to improvement in the baseline model and only in the case of $\alpha=0$ where only the likelihood cost function was in use. Note that in this case the training step is only based on the labeled data.
In all the other cases the post-processing did not help.

\subsection{Analysis}

The ladder model exhibited superior regularization, as was evident in the cross validation tests we made.
Table \ref{restable} shows the cross validation performance results for the ladder and the baseline model (without the ladder reconstruction score).
The score was defined in (\ref{evalcost}) where the number of languages was set to be $k=38$.
Each row of the table corresponds to a different split of the 50 languages into 38 in-set and 12 out-of-set languages.
Selecting different sets of languages to be in-set and
out-of-set can yield  drastically different results. We therefore
repeated the process 5 times to allow for each language
 to be out-of-set at least once.

Table \ref{restable} also reports the  number of epochs that were needed  to get the best test results.
In the baseline method there was a problem of over-fitting and that required  early stopping
to achieve best results.  By contrast, the ladder score imposes an aggressive regularization which avoids the need for early stopping.
Figure  \ref{fig_Ladder_score} shows the cost function that was minimized during training as a function of the
number of epochs. The cost function is plotted for both  the training data and the test data.
It is clear  that, while in the baseline method
 the problem of overfitting could not be avoided, the Ladder Network has no overfitting problems.
This behavior is yet another advantage of the Ladder Network training procedure in that we can train the network
up to  local optimum without the risk of overfitting.

To conclude, in this study we utilized the Ladder Network technique for semi-supervised learning of a language identification system.
We showed that the Ladder Network provided an improved framework for integrating unlabeled data into a supervised training scheme.
The Ladder Network also uses  the unlabeled data in the cost function optimized at the training step. This enabled us to explicitly address out-of-set examples in part of the training process. We also used a new cost function that encourages the statistics of the predicted labels in the unlabeled dataset to be similar to the label statistics of the labeled training data. The joint contribution of the ladder training methods and this cost function provided a significant reduction in classification errors  of 15\% (from  28.41 to 23.95) when applied to the  NIST 2015 Language Recognition Challenge.

{\small
\bibliographystyle{IEEEbib}
\bibliography{paper}
}

    \end{document}